\documentclass[10pt,twocolumn,letterpaper]{article}
\usepackage{wacv}
\usepackage{times}
\usepackage{epsfig}
\usepackage{graphicx}
\usepackage{amsmath}
\usepackage{amssymb}
\usepackage{amsfonts}
\usepackage[numbers,sort&compress]{natbib}
\usepackage{float}
\usepackage{bm}
\usepackage{multirow}
\usepackage{multicol}

\usepackage[pagebackref=true,breaklinks=true,letterpaper=true,colorlinks,bookmarks=false]{hyperref}

\wacvfinalcopy 

\ifwacvfinal\fi

\author{Yunxiao Shi$^{1,2}$ \hspace{0.5cm} Jing Zhu$^{1,2,3}$ \hspace{0.5cm} Yi Fang$^{1,2,3}$\thanks{indicates corresponding author.} \hspace{0.5cm} Kuochin Lien$^{4}$ \hspace{0.5cm} Junli Gu$^{4}$\vspace{0.2cm}\\
	${}^1$NYU Multimedia and Visual Computing Lab, USA \\
	${}^2$New York University, USA \\
	${}^3$New York University Abu Dhabi, UAE\\
	${}^4$XMotors.ai, USA\\
	{\tt\small \{yunxiao.shi, jingzhu, yfang\}@nyu.edu \{kuochin,junli\}@xmotors.ai}}
\begin{document}

\title{Self-Supervised Learning of Depth and Ego-motion with Differentiable Bundle Adjustment}


\maketitle

\begin{abstract}
    
    Learning to predict scene depth and camera motion from RGB inputs only is a challenging task. Most existing learning based methods deal with this task in a supervised manner which require ground-truth data that is expensive to acquire. More recent approaches explore the possibility of estimating scene depth and camera pose in a self-supervised learning framework. Despite encouraging results are shown, current methods either learn from monocular videos for depth and pose and typically do so without enforcing multi-view geometry constraints between scene structure and camera motion, or require stereo sequences as input where the ground-truth between-frame motion parameters need to be known. In this paper we propose to jointly optimize the scene depth and camera motion via incorporating differentiable Bundle Adjustment (BA) layer by minimizing the feature-metric error, and then form the photometric consistency loss with view synthesis as the final supervisory signal. The proposed approach only needs unlabeled monocular videos as input, and extensive experiments on the KITTI and Cityscapes dataset show that our method achieves state-of-the-art results in self-supervised approaches using monocular videos as input, and even gains advantage to the line of methods that learns from calibrated stereo sequences (i.e. with pose supervision). 
\end{abstract}

\section{Introduction}

Humans are remarkably capable of inferring ego-motion and 3D structure of an unseen scene even with a single glance. Years of computer vision research has been trying to equip computers with similar modeling capabilities but is still far away from human-level performance. Therefore, 
\begin{figure}[H]
	\begin{center}
		\includegraphics[width=1.0\linewidth]{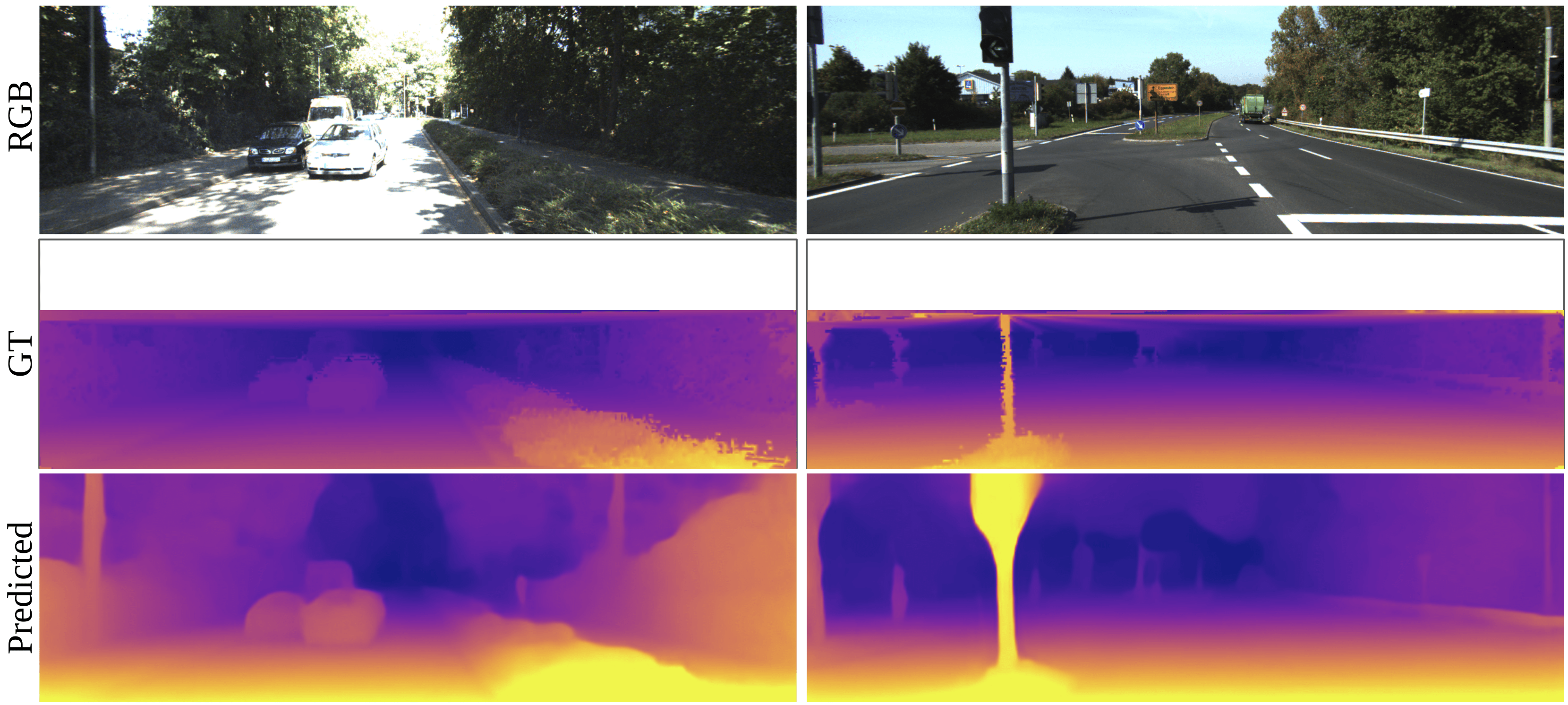}
	\end{center}
	\caption{Our example depth prediction results on KITTI 2015. The top row is input RGB image, the middle is ground truth disparities, and the last row is our prediction. Our method is capable of capturing thin structures like poles and street signs.}
	\label{fig:long}
	\label{fig:onecol}
\end{figure}
\noindent understanding 3D scene structure from a single image (i.e. monocular depth estimation) and inferring ego-motion from a sequence of images (i.e. camera pose estimation) remains two fundamental problems in machine perception. These two problems are crucial in robotic vision research since accurate image-based depth estimation and ego-motion inference is of vital importance for many applications, among which most notably for autonomous vehicles. 

Researchers have been devoting lots of efforts to both problems since the earliest days of machine vision research, with many geometry based methods proposed early on \cite{mur2015orb, engel2014lsd, engel2018direct}. Recently with the impressive performance deep learning powered methods have demonstrated in various 2D and 3D computer vision tasks \cite{he2016deep, long2015fully, ren2015faster, qi2017pointnet, zhou2018voxelnet}, a number of works have cast the problem of depth estimation and motion inference as supervised learning tasks \cite{eigen2014depth, li2015depth, liu2015deep, wang2015towards}. These methods attempt to predict depth and camera pose using models that have been trained on a large dataset consisting of various real-world scenes. However, the ground truth depth or pose annotations are expensive to obtain (e.g. high-end LIDAR or depth camera to collect depth). Recently Garg \textit{et al.} \cite{garg2016unsupervised} realized first that the depth estimation is feasible in an unsupervised learning framework and proposed to use photometric warp error as a self-supervisory signal to train a convolutional neural network (CNN) to predict scene depth. Zhou \textit{et al.} \cite{zhou2017unsupervised} explored the possibility to only use monocular videos as input and jointly trained two networks to estimate depth and pose. Others \cite{wang2018learning, yang2018unsupervised, luo2018every} made improvements by taking optical flow or moving objects into account. However, although the results are encouraging, there still exists a gap between training from stereo and monocular data.

In this work, we propose to learn scene structure and camera motion from monocular videos in an self-supervised manner with differentiable Bundle Adjustment (BA). The differentiable BA is formulated as a feature-metric BA layer as in \cite{tang2018ba} but with modifications, which takes CNN features from multiple images as inputs and optimizes for the scene structure and camera motion by minimizing a feature-metric error. Multi-view geometric constraints between 3D scene structures and camera motion are enforced via this feature-metric BA component in our network. In contrast to geometric BA which only utilizes a limited portion of image information (e.g. image edges or corners, blobs, etc), or photometric BA that is sensitive to camera exposure or moving objects, the feature-metric BA can back-propagate loss from scene structure and camera motion to learn most suitable features. In order to make the feature-metric BA differentiable with respect to input CNN features, the \textit{if-else} logics (number of iterations and the value of damping factor $\lambda$) in the original Levenberg-Marquardt algorithm \cite{nocedal2006numerical} that is ubiquitously used to solve BA has to be removed or relaxed. To this end we fix the number of iterations referred to as `incomplete optimization' in \cite{domke2012generic} and send the current feature-metric error to a Multi-Layer Perceptron (MLP) to predict the damping factor $\lambda$. Then we use the output scene structure (represented by per-pixel depths) and camera motion (represented by $4\times 4$ transformation matrices) to synthesize views to construct the photometric consistency loss. New state-of-the-art performance in self-supervised approaches using monocular videos as input are achieved. Our approach also compares competitively to self-supervised methods learning from stereo pairs.

\begin{figure*}
	\begin{center}
		\includegraphics[width=1.0\textwidth]{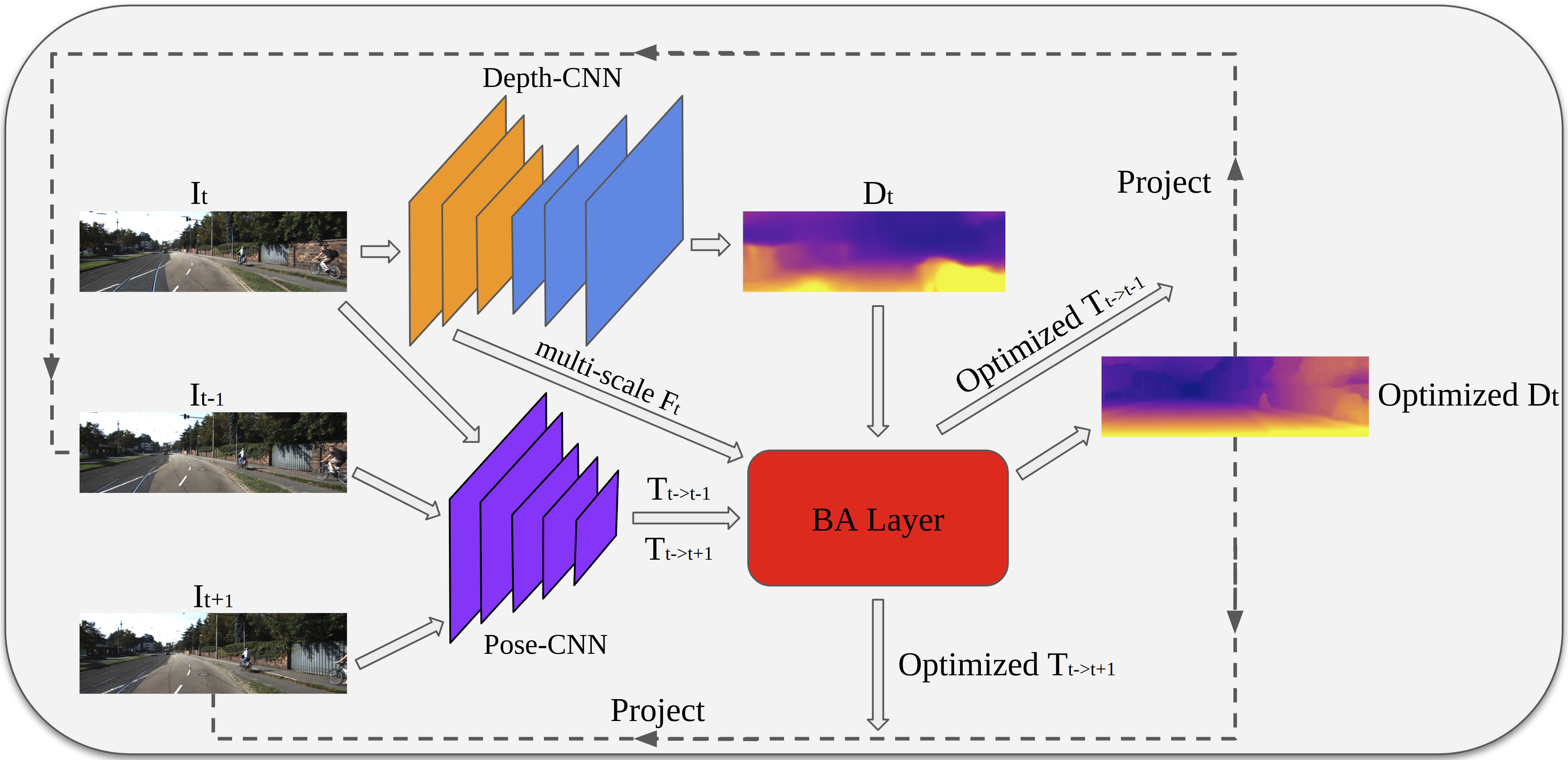}
	\end{center}
	\caption{Pipeline of our system. Our network takes a sequence of images (here we use a 3-frame-long tracklet for illustration) as input. A dense depth map $D_t$ is first generated for the target view $I_t$ by passing it through a depth-CNN. Meanwhile we construct feature pyramids $F_t$, $F_{t-1}$, $F_{t+1}$ for both $I_t$ and its adjacent two source views $I_{t-1}$ and $I_{t+1}$ using the same encoder part in the depth-CNN (shown in Fig. \ref{fig:fpn}). The pose-CNN takes a pair of images as input and outputs relative poses $T_{t\rightarrow t-1}$ and $T_{t\rightarrow t+1}$. The image features $F_{\{t,t-1,t+1\}}$ (only $F_t$ are shown in this figure due to space), depth map $D_t$ (downsampled to corresponding scales), relative poses $T_{t\rightarrow t-1}$ and $T_{t\rightarrow t+1}$ are fed into the BA layer (see Fig. \ref{fig:balayer}) to jointly optimize scene depth and camera pose.  Then we can reconstruct $I_t$ with $I_{t-1}$ and $I_{t+1}$ using differentiable image warping \cite{jaderberg2015spatial} to form photometric consistency loss to train the system.}
	\label{fig:model}
\end{figure*}

\section{Related Work}

\subsection{Supervised Depth and Ego-Motion Learning} 
Before the rise of deep learning based methods, previous approaches have tried to estimate depth from a single image based on hand-crafted features \cite{hoiem2005automatic}, Markov Random Fields \cite{saxena2006learning, saxena2009make3d} or semantic segmentation \cite{ladicky2014pulling}. Eigen \textit{et al.} \cite{eigen2014depth} is the first to use ConvNets to predict depth. They proposed a multi-scale approach with two CNNs for depth estimation,where a coarse-scale network first predicts scene depth at global level and then a fine-scale network refines the local regions. Liu \textit{et al.} \cite{liu2015deep, liu2016learning} utilized a deep convolutional neural field model to estimate depth from single monocular images. Laina \textit{et al.} \cite{laina2016deeper} proposed using fully convolutional residual network coupled with reverse Huber loss to predict depth. Kendall \textit{et al.} \cite{kendall2017end} proposed an end-to-end learning pipeline to perform stereo matching. Xu \textit{et al.} \cite{xu2017multi} constructed multi-scale CRFs for depth prediction. For ego-motion learning, Agrawal \textit{et al.} \cite{agrawal2015learning} proposed an algorithm that uses ego-motion as supervision to learn good visual features and it is capable of estimating relative camera poses. Wang \textit{et al.} \cite{wang2017deepvo} proposed a recurrent ConvNet architecture for learning monocular odometry from video sequences. Ummenhofer \textit{et al.} \cite{ummenhofer2017demon} presented a depth and motion network where the scene depth and camera motion are predicted from optical flow. In addition to depth and pose ground truth, \cite{ummenhofer2017demon} also needs annotations of surface normal and optical flow between images. 

\subsection{Self-Supervised or Semi-Supervised Approaches}
The assumption that adjacent frames should be photometrically consistent sparkled recent works to learn scene depth in an unsupervised pipeline. There are two categories in the pursuit of this direction. The first category learns depth from stereo sequences, where the between-frame camera poses are known. Garg \textit{et al.} \cite{garg2016unsupervised} used stereo pairs and trained a network which minimizes the photometric distance between the true right image and the synthesized image which is warped from the left image using the estimated depth. Godard \textit{et al.} \cite{godard2017unsupervised} improved upon \cite{garg2016unsupervised} by introducing a better stereo loss function and also a left-right disparity consistency which enforced geometric constraints during training. Besides the left-right photometric error, the temporal photometric and feature reconstruction errors were also considered to improve performance \cite{zhan2018unsupervised}. There are also recent attempts \cite{aleotti2018generative,pilzer2018unsupervised} to exploit the use GANs \cite{goodfellow2014generative} to better benefit unsupervised depth learning.

The second category is to learn depth from monocular video sequences. Compared to the first category this is more challenging since the between-frame camera motion are unknown. Nonetheless recently Zhou \textit{et al.} \cite{zhou2017unsupervised} and Vijayanarasimhan \textit{et al.} \cite{vijayanarasimhan2017sfm} showed that it is possible to learn scene depth and camera motion simultaneously using the supervisory signal from photometric consistency error and spatial smoothness loss. \cite{zhou2017unsupervised} also used a motion explanation mask to deal with regions that violated the rigid-scene assumption but later on experiments showed improved performance after disabling this term. The pose estimation network removes the need of stereo image pairs as training samples. Several more recent works took into account different constraints in training to improve upon the above two works. Since predicting depth and optical flow are related tasks, \cite{yin2018geonet, zou2018df} demonstrated that jointly learning depth and optical flow can benefit each other. Inspired by the direct visual odometry techniques \cite{engel2014lsd, engel2018direct}, Wang \textit{et al.} \cite{wang2018learning} introduced a differentiable module for camera pose estimation to replace the previous pose estimation network after depth is estimated by the method in \cite{zhou2017unsupervised}. They also proposed a depth normalization strategy to overcome the scale sensitivity problem. 

\section{Our Approach}
This section details our approach to learn scene depth and camera motion with differentiable BA layer from unlabeled monocular videos. An auto-encoder depth CNN is used to generate dense per-pixel depth map for the target view image. The encoder component in the depth CNN is used (i.e. shared weights) to construct image feature pyramids for target view image and two adjacent source view images. The pose CNN takes a pair of images as input to output relative camera poses. The BA layer takes the image features, depth map (downsampled to corresponding scales if needed) and relative poses to jointly optimize scene structure and camera motion via a feature-metric error. We then use the predicted depth and pose to construct the photometric consistency loss as the supervisory signal to train our model.
 
\subsection{Image Feature Pyramid Construction} \label{sec31}
\begin{figure}
 	\begin{center}
 		\includegraphics[width=0.5\textwidth]{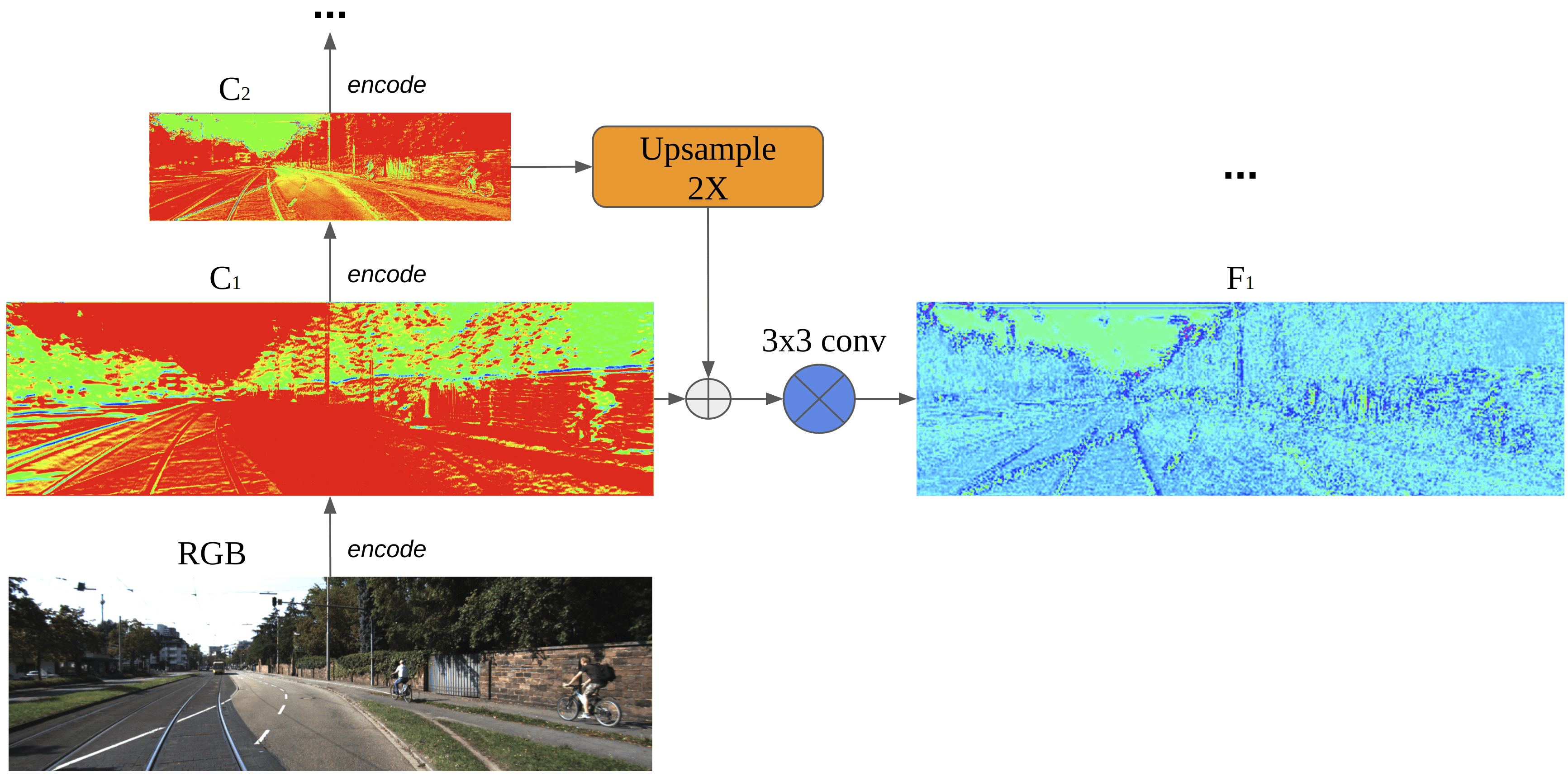}
 	\end{center}
 	\caption{Construction of multi-scale feature maps for an input RGB image. Here we show the generation of $F_1$ as illustration.}
 	\label{fig:fpn}
 \end{figure}

 Here we exploit the inherent multi-scale hierarchy of deep convolutional networks to construct feature pyramids to learn better features for the downstream feature-metric BA. A top-down architecture with lateral connections is developed to extract high-level semantic information from coarse to fine scales. Here we use the fully-convolutional U-Net \cite{ronneberger2015u} architecture as the backbone. As shown in Figure \ref{fig:fpn}, an input image is fed into the U-Net encoder to extract feature maps at different scales. We denote the resulting feature map as $C_k,k\in\{1,2,3,4\}$. Then we upsample the feature map $C_{j+1},j\in\{1,2,3\}$ with bilinear interpolation by a scale factor of 2 and concatenate it with $C_j$.  A $3\times 3$ convolution is then applied to the concatenated feature maps. The final feature pyramid for an input image $I_i$ is $\bm{F}_i = \{F_{i_1}, F_{i_2}, F_{i_3}\}$. 

\subsection{Depth Representation}
Dense depth plays a critical role in many tasks such as 3D object detection and robot navigation. Here we use the standard per-pixel depth representation which is the most common and demonstrated to work well in self-supervised learning settings \cite{zhou2017unsupervised,godard2017unsupervised}. As shown in Fig. \ref{fig:model}, we train U-Net as our depth CNN to generate a dense depth map $D_i$ for an input image $I_i$ as in Eq. \ref{eq1}, 
\begin{equation}
    D_i = f(I_i),
    \label{eq1}
\end{equation}
where and $f(\cdot)$ is our depth CNN. The generated depth $D_i$ is the same resolution as the original input image, and will be downsampled accordingly to couple with the image feature maps of different scales as input to the downstream Bundle Adjustment layer for further optimization. 

\subsection{Pose Network}

For pose estimation we follow previous works \cite{zhou2017unsupervised,wang2018learning} and predict the rotation part in the camera pose using Euler angle representation, and also the predicted translation is divided by the estimated depth mean to align the scale of translation to depth prediction. We designed our pose network based on ResNet-18 \cite{he2016deep} but modified it to take a pair of RGB images as input and to out a 6-DoF relative pose as shown in Fig. \ref{fig:model}. We note that with the predicted 6-DoF relative poses we can easily convert them into $4\times 4$ transformation matrices by the exponential mapping between $\mathfrak{se}(3)$ and $SE(3)$ for later-on re-projection.

\subsection{Bundle Adjustment Layer} \label{sec33}
We first quickly recap the classic formulations of bundle adjustment. Let $\bm{I}=\{I_i\}_{i=1}^N$ denote a sequence of images, the geometric BA \cite{triggs1999bundle, agarwal2010bundle} jointly optimizes camera pose $\bm{\mathrm{T}} = \{T_i\}_{i=1}^N$ and scene point coordinates $\bm{\mathrm{P}}=\{P_j\}_{j=1}^M$ by minimizing the re-projection error,
\begin{equation}
    \mathcal{S} = \arg\min\sum_{i=1}^N\sum_{j=1}^Me_{i,j}^{geo}(\mathcal{S}),
    \label{eq2}
\end{equation}
where the geometric distance
\begin{equation*}
    e_{i,j}^{geo} = ||\pi(T_i, P_j) - q_{i, j}||_2^2
\end{equation*}
measures the difference between the observed feature point and the projected 3D scene point, with $\pi(\cdot)$ being the projection and $||\cdot||_2$ the $\ell_2$ norm. $q_{i,j} = (x_{i,j}, y_{i, j}, 1)^{\top}$ is the normalized homogeneous pixel coordinate. $\mathcal{S} = \{T_1, \cdots, T_N, P_1, \cdots, P_M\}^{\top}$ contains all the parameters. To minimize Eq. \ref{eq2} the Levenberg-Marquardt (LM) \cite{nocedal2006numerical} is used as the general strategy. The LM at each iteration solves for an optimal update $\Delta\mathcal{S}^{*}$
\begin{equation}
    \Delta\mathcal{S}^{*} = \arg\min||E(\mathcal{S})+J(\mathcal{S})\Delta(\mathcal{S})||_2^2+\lambda||D(\mathcal{S})\Delta\mathcal{S})||_2^2,
\end{equation}
where $E(\mathcal{S})=\{e_{1,1}^{geo}(\mathcal{S}), e_{1,2}^{geo}(\mathcal{S}), \cdots, e_{N, M}^{geo}(\mathcal{S})\}$, and $J(\mathcal{S})$ is the Jacobian matrix of $E(\mathcal{S})$ with respect to $\mathcal{S}$, $D(\mathcal{S})$ is a non-negative diagonal matrix, typically being the square root of the diagonal of the approximated Hessian $J(\mathcal{S})^{\top}J(\mathcal{S})$, and $\lambda$ being the damping factor.
\begin{figure*}
	\begin{center}
		\includegraphics[width=1.0\textwidth]{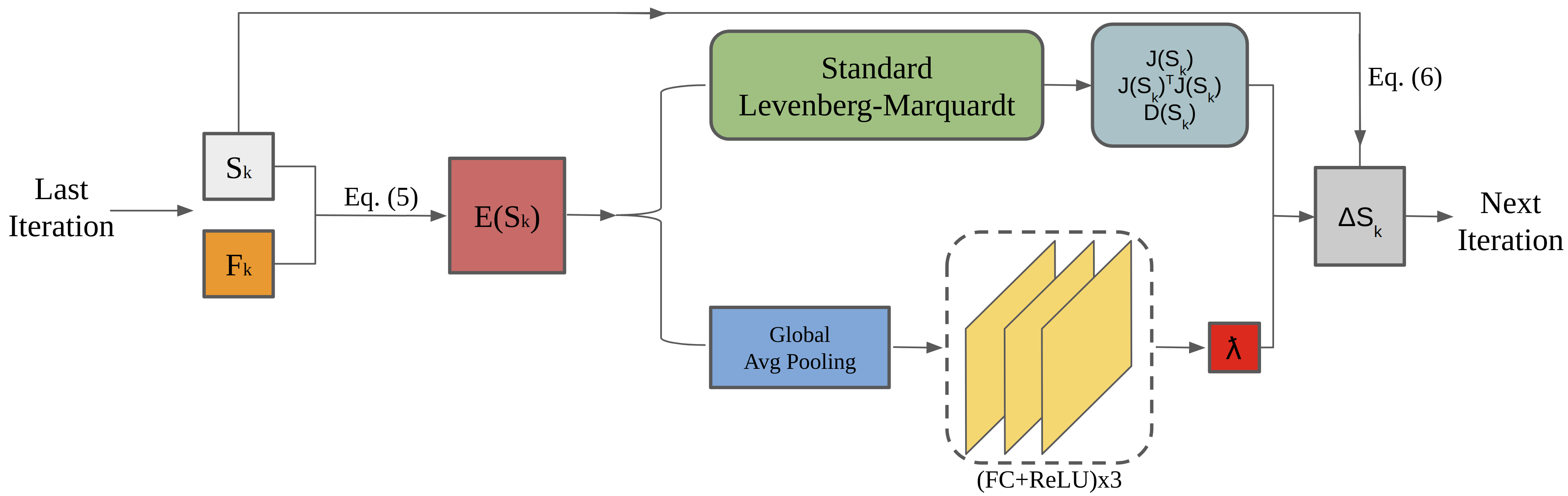}
	\end{center}
	\caption{Illustration of a single step in differentiable Levenberg-Marquardt as in \cite{tang2018ba}. Given solution $\mathcal{S}_k$ in the previous iteration and input image features, we first compute feature-metric error $E(\mathcal{S}_k)$ by Eq. \ref{eq5} and the Jacobian $J(\mathcal{S}_k)$ with respect to $\mathcal{S}_k$. After that we can compute the approximated Hessian $J(\mathcal{S}_k)^{\top}J(\mathcal{S}_k)$ and the diagonal matrix $D(\mathcal{S}_k)$. Meanwhile we apply global average pooling to $E(\mathcal{S}_k)$ to get a fixed-length vector and pass it through three stacked fully connected layers to predict $\lambda$. We use ReLU activation to ensure that $\lambda$ will be non-negative. The update $\Delta\mathcal{S}_k$ can then be computed and sent into next iteration.}
	\label{fig:balayer}
\end{figure*}

As previously pointed out the geometric BA only uses a limited portion of image information, and requires good feature point matching to work well, which typically can not be guaranteed even with outlier rejection criteria like RANSAC \cite{fischler1981random}. To overcome these limitations, the development of direct methods \cite{engel2014lsd, engel2018direct} proposed photometric BA to eliminate feature matching and minimize the photometric error of pixel intensity difference
\begin{equation}
	e_{i, j}^{photo}(\mathcal{S}) = I_i(\pi(T_i, d_j\cdot q_j)) - I_1(q_j),
\end{equation}
where $d_j$ is the depth of a pixel $q_j$ at image $I_1$, and $\mathcal{S}=\{T_1, \cdots, T_{N}, d_1, \cdots, d_N\}^{\top}$. The direct methods have the advantage of potentially using all pixels and have demonstrated superior performance compared to geometry based methods. But as pointed out in \cite{mur2015orb, engel2018direct} they also have the drawbacks of being sensitive to initialization and camera exposure. 

To deal with the above challenges, we use the feature-metric error proposed in \cite{tang2018ba} to jointly optimize scene depth and camera motion, which minimizes the feature-metric error, 
\begin{equation}
    e_{i,j}^{feat}(\mathcal{S}) = F_i(\pi(T_i, d_j\cdot q_j)) - F_1(q_j),
    \label{eq5}
\end{equation}
where $F_i$ are feature pyramids which are constructed as in section \ref{sec31} of images $I_i$, and the optimization parameters $\mathcal{S}$ are the same as in photometric BA. The feature-metric utilizes more image information than only corners or blobs, and meanwhile is more robust to initialization or exposure changes which is desirable for learning depth and pose from images. 

In order to back-propagate the loss information, the solution $\mathcal{S}$ should be differentiable with respect to the image features $F$. In the original Levenberg-Marquardt algorithm the number of iterations is determined using a threshold rule and the value of damping factor $\lambda$ is increased if an optimization step fails to reduce the error and decreased otherwise. These two \textit{if-else} logics makes it impossible to differentiate the loss with respect to the image features therefore have to be softened or removed. We fix the number of iterations to solve the first issue which also reduces memory consumption. For the value of $\lambda$, as shown in Fig. \ref{fig:balayer}, we use the same architecture of the BA layer in \cite{tang2018ba}. By this formulation each LM step becomes differentiable and loss information can be back-propagated.

\subsection{View Synthesis as Supervision} \label{sec34}
Given a target view image $I_t$ and source view image $I_s$, with the predicted depth $\hat{D}_t$ and relative pose $\hat{T}_{t\rightarrow s}$ in section \ref{sec33}, we can reconstruct the target view $\hat{I}_s$. We denote $p_t$ the homogeneous coordinate of a pixel in the target view, and $K$ the camera intrinsics. The projected coordinate $p_s$ on the source view can be obtained as 
\begin{equation}
	p_s\sim K\hat{T}_{t\rightarrow s}\hat{D}_t(p_t)K^{-1}p_t.
\end{equation}
Such obtained $p_s$ are continuous values. To obtain $\hat{I}_s(p_t)$ here we follow \cite{zhou2017unsupervised} using the differentiable bilinear sampling mechanism proposed in \cite{jaderberg2015spatial} to linearly sample the four neighbourhood pixels, i.e. 
\begin{equation}
\hat{I}_s(p_t) = \sum_i\sum_jw^{ij}I_s(p_s^{ij}),
\end{equation}
where $i, j$ represents the top-bottom, left-right direction respectively, and $\sum_{i,j}w_{ij}=1$ are weights linearly proportional to the Euclidean distance between $p_s$ and $p_s^{ij}$.

\subsection{Training Loss}
With synthesized view $\hat{I}_s$ corresponding to $I_t$ constructed in section \ref{sec34}, the photometric consistency loss can be formulated as 
\begin{equation}
	\mathcal{L}_{photo} = \sum_s\sum_p |I_t(p) - \hat{I}_s(p)|,
\end{equation}
where $p$ is all image pixels and $s$ index over all source views. To encourage predicted depth to be locally smooth, we adopt the inverse depth smoothness loss \cite{heise2013pm,godard2017unsupervised}
\begin{equation}
    \mathcal{L}_s = \sum_{i}\sum_j|\partial_xD_{ij}|e^{-|\partial_xI_{ij}|} + |\partial_yD_{ij}|e^{-|\partial_yI_{ij}|},
\end{equation}
where $\partial_x(\cdot)$ and $\partial_y(\cdot)$ are image gradients in the horizontal and vertical directions respectively. Furthermore we pose constraint on the warped $\hat{I}_s$ to be structurally consistent with $I_t$ with the single scale SSIM \cite{wang2004image} to incorporate the appearance matching loss \cite{godard2017unsupervised} into the training objective
\begin{equation}
    \mathcal{L}_{m} = \sum_i\sum_j\alpha\frac{1-\text{SSIM}(I_t,\hat{I}_s)}{2} + (1-\alpha)\mathcal{L}_{photo}.
    \label{eq11}
\end{equation}
Therefore the final training loss is 
\begin{equation}
    \mathcal{L} = \mathcal{L}_m + \lambda_s\mathcal{L}_s,
    \label{eq12}
\end{equation}
where $\alpha$ and $\lambda_s$ are scalars to balance different loss terms.

\section{Experiments}
Here we test the performance of our approach and compare with prior works on monocular depth and ego-motion estimation. We train on unlabeled monocular videos which have neither depth nor pose ground truth supervision. We use the KITTI \cite{geiger2013vision} and Cityscapes \cite{cordts2016cityscapes} dataset to evaluate our system.

\vspace{5pt}
\noindent\textbf{Implementation Details} We implemented our model using the publicly available framework PyTorch \cite{paszke2017automatic}. For all experiments, we set $\alpha=0.85$ in Eq. \ref{eq11} and $\lambda_s=0.1/r$ in Eq. \ref{eq12}, where $r$ is the corresponding downscale ratio with respect to the input image. We solve the feature-metric BA on the respective 3 scales with feature map warping, and set the number of iterations to 6 on each level following \cite{engel2018direct}, leading to 18 iterations in total. Batch normalization \cite{ioffe2015batch} is used for all layers except the output layers. We use the Adam optimizer \cite{kingma2014adam} with $\beta_1=0.9$, $\beta_2=0.999$, $\epsilon=10^{-8}$, an initial learning rate of $0.0002$ and a mini-batchsize of 4. The number of hidden units in the MLP to predict the damp factor $\lambda$ in Levenberg-Marquardt is set to 128. The images are resized to have resolution of $192\times 640$ in all experiments. Data augmentation with a 0.5 probability of horizontal flip, gamma, brightness and color adjustments are performed on the fly. The values are sampled from uniform distributions in the ranges of $[0.8, 1.2]$ for gamma, $[0.5, 2.0]$ for brightness and $[0.8, 1.2]$ per each color channel separately following \cite{godard2017unsupervised}. The learning rate is decayed in half once we observe the training plateaus. The corresponding parameter update are addition for depth and $\mathfrak{se}(3)$ exponential mapping for camera pose. Due to the computation heavy differentiable BA optimization the training typically takes about 60 hours on a NIVDIA TITAN Xp GPU. 
\begin{figure*}
	\begin{center}
		\includegraphics[width=1.0\textwidth]{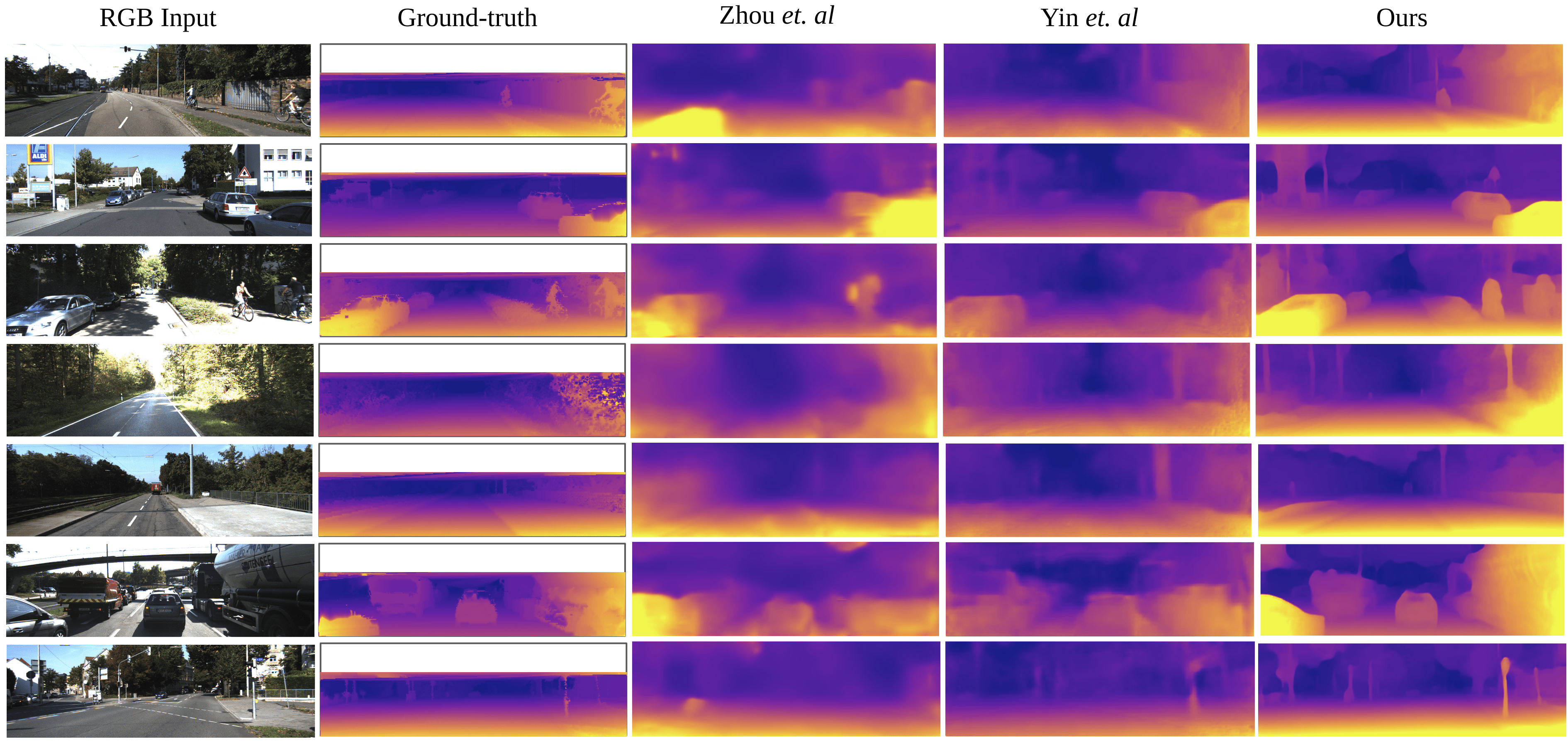}
	\end{center}
	\caption{Comparison of single-view depth estimation results on the KITTI Eigen \textit{et al.} \cite{eigen2014depth} test split. The ground velodyne depths are sparse therefore we interpolate them for visualization. The first and second column are input RGB images and ground-truth depth maps respectively. The rest are depth predictions from various methods with ours at the rightmost position. Best viewed in color.}
	\label{fig:kittiquality}
\end{figure*}

\subsection{Monocular Depth Estimation}
\subsubsection{KITTI Dataset} 
The KITTI dataset \cite{geiger2013vision} is a widely used benchmark dataset composed of several outdoor scenes captured while driving with car-mounted cameras and a LIDAR sensor. It contains 61 scenes in total belonging to one of the three categories: ``city'', ``residential'' and ``road''. We train our system on the same data split provided by Eigen \textit{et al.} \cite{eigen2014depth} which covers 29 scenes and we use the remaining 32 scenes for training. We exclude all the static frames whose mean optical flow magnitude is too small since too many static frames will cause the network to always output identity pose and consequently depth estimation will also diverge. Images captured by both cameras are used but are treated independently when forming training sequences as in \cite{zhou2017unsupervised}. This results in a total of 44,372 sequences, out of which we use 40,378 for training and 3,994 for validation. We set the length of image sequences to be 3, and treat the central frame as target view, and the temporally adjacent previous and next frames as source views.  By the formulation of our model and the monocular setting, the depth predicted by our model is up to a certain scale. We align the scale of our predictions to the ground-truth by multiplying the former with a scalar $\hat{s} = $\textit{median}($D_{gt}$)/\textit{median}($D_{pred}$) to make sure the comparison is fair with previous works.

We evaluate the single-image depth estimation performance on the 697 images from the test split of Eigen \textit{et al} \cite{eigen2014depth}. For evaluation metrics, we evaluate our method using several errors from prior works \cite{eigen2014depth, liu2016learning}:
\begin{align}
    &\text{Threshold: percentage of }y_i\,\, s.t. \max(\frac{y_i}{y_i^{*}}, \frac{y_i^{*}}{y_i})=\delta<\theta,\notag\\
    &\text{Abs Relative difference: }\frac{1}{|T|}\sum_{y\in T}\frac{|y-y^{*}|}{y^{*}},\notag\\
    &\text{Squared Relative difference: } \frac{1}{|T|}\sum_{y\in T}\frac{||y-y^{*}||^2}{y^{*}},\notag\\
    &\text{RMSE (linear)}: \sqrt{\frac{1}{|T|}\sum_{y\in T}||y_i - y_i^{*}||^2},\notag\\
    &\text{RMSE (log)}: \sqrt{\frac{1}{|T|}\sum_{y\in T}||\log y_i - \log y_i^{*}||^2},\notag
\end{align}
where $\theta$ is the threshold which is set to $1.25$ following all previous works. $y_i$ is the predicted depth and $y_i^{*}$ is the ground-truth. As shown in Table \ref{table1}, our method out-performs all recent unsupervised approaches using monocular videos as input (e.g. Yin \textit{et al}. \cite{yin2018geonet} and Wang \textit{et al.} \cite{wang2018learning}), and even gains advantage compared to methods trained using calibrated stereo image pairs (i.e. with pose supervision) (e.g. Godard \textit{et al.} \cite{godard2017unsupervised}). Figure \ref{fig:kittiquality} provides example visualizations of our method and several prior approaches on the KITTI Eigen test split \cite{eigen2014depth}. We note that since the ground truth velodyne depths are sparse therefore we interpolate to them for visualization and cropped out the top part which are invalid measurements.  One can see that our method better preserves thin structures such as street lights and trees than previous methods which indicates the effectiveness of the proposed method.
\begin{table*}[h]
	\centering
	\footnotesize
	\begin{tabular}{p{2.5cm}p{0.8cm}p{1cm}p{1.2cm}p{1cm}p{1cm}p{1.5cm}p{1.2cm}p{1.5cm}p{1.5cm}}
		\hline\hline
		Method & Dataset & Supervision & \multicolumn{4}{c}{Error Metric} & \multicolumn{3}{c}{Accuracy Metric}\\
		 & & & \centering Abs Rel &\centering Sq Rel &\centering RMSE &\centering RMSE (log) & $\delta<1.25$& $\delta<1.25^2$ & $\delta<1.25^3$\\\hline
		 Eigen \textit{et al.} \cite{eigen2014depth} coarse & \centering K &\centering depth &\centering 0.214 &\centering 1.605 &\centering 6.563 &\centering 0.292 & \centering 0.673 &\centering 0.884 & \centerline{0.957}\\[-2.5ex]
		 Eigen \textit{et al.} \cite{eigen2014depth} fine &\centering K &\centering depth &\centering 0.203 &\centering 1.548 &\centering 6.307 &\centering 0.282 & \centering 0.702 & \centering 0.890 &\centerline{0.958}\\[-2.5ex]
		 Liu \textit{et al.} \cite{liu2016learning} & \centering K & \centering depth & \centering 0.202 &\centering 1.614 &\centering 6.523 &\centering 0.275 & \centering 0.678 & \centering 0.895 &\centerline{0.965}\\[-2.5ex]
		 \text{Garg \textit{et al.}} \cite{garg2016unsupervised} & \centering K & \centering pose & \centering 0.169 & \centering 1.080 & \centering 5.104 & \centering 0.273 & \centering 0.740 & \centering 0.904 & \centerline{0.962}\\[-2.5ex]
		 Godard \textit{et al.} \cite{godard2017unsupervised} & \centering K & \centering pose & \centering 0.148 & \centering 1.344 & \centering 5.927 & \centering 0.247 & \centering 0.803 & \centering 0.922 & \centerline{0.964}\\[-2.5ex]
		 Ramirez \textit{et. al} \cite{ramirez2018geometry} & \centering K & \centering pose+category & \centering 0.143 & \centering 2.161 & \centering 6.127 & \centering 0.210 & \centering 0.854 & \centering 0.945 & \centerline{0.976}\\[-2.5ex]
		 Yang \textit{et. al} \cite{yang2018deep} & \centering K & \centering pose & \centering 0.097 & \centering 0.734 & \centering 4.442 & \centering 0.187 & \centering 0.888 & \centering 0.958 & \centerline{0.980}\\[-2.5ex]
		 Fu \textit{et. al} \cite{fu2018deep} & \centering K & \centering depth & \centering \textbf{0.072} & \centering \textbf{0.307} & \centering \textbf{2.727} & \centering \textbf{0.120} & \centering \textbf{0.932} & \centering \textbf{0.984} & \centerline{\textbf{0.994}}\\[-2.5ex]\hline
		 Zhou \textit{et al.} \cite{zhou2017unsupervised} & \centering K & \centering none & \centering 0.208 & \centering 1.768 & \centering 6.856 & \centering 0.283 & \centering 0.678 & \centering 0.885 & \centerline{0.957}\\[-2.5ex]
		 Zhou \textit{et al.} \cite{zhou2017unsupervised}$^{*}$ & \centering K & \centering none & \centering 0.183 & \centering 1.595 & \centering 6.709 & \centering 0.270 & \centering 0.734 & \centering 0.902 & \centerline{0.959}\\[-2.5ex]
		 Yin \textit{et al.} \cite{yin2018geonet} & \centering K & \centering none & \centering 0.155 & \centering 1.296 & \centering 5.857 & \centering 0.233 & \centering 0.793 & \centering 0.931 & \centerline{0.973}\\[-2.5ex]
		 Wang \textit{et al.} \cite{wang2018learning}& \centering K & \centering none & \centering 0.151 & \centering 1.257 & \centering 5.583 & \centering 0.228 & \centering 0.810 & \centering 0.936 & \centerline{0.974}\\[-2.5ex]
		 Poggi \textit{et. al} \cite{poggi2018learning} & \centering K & \centering none & \centering 0.114 & \centering 1.088 & \centering 5.756 & \centering \textbf{0.203} & \centering 0.848 & \centering 0.944 & \centerline{\textbf{0.979}}\\[-2.5ex]
		 Poggi \textit{et. al} \cite{poggi2018towards} & \centering K & \centering none & \centering 0.153 & \centering 1.363 & \centering 6.030 & \centering 0.252 & \centering 0.789 & \centering 0.918 & \centerline{0.963}\\[-2.5ex]
		 Zou \textit{et. al} \cite{zou2018df} & \centering K & \centering none & \centering 0.150 & \centering 1.124 & \centering 5.507 & \centering 0.223 & \centering 0.806 & \centering 0.933 & \centerline{0.973}\\[-2.5ex]
		 Mahjourian \textit{et. al} \cite{mahjourian2018unsupervised} & \centering K & \centering none & \centering 0.163 & \centering 1.240 & \centering 6.220 & \centering 0.250 & \centering 0.762 & \centering 0.916 & \centerline{0.968}\\[-2.5ex]\hline
		 Ours (diff-BA) & \centering K & \centering none & \centering \textbf{0.113} & \centering \textbf{1.079} & \centering \textbf{\textbf{4.931}} & \centering 0.206 & \centering \textbf{0.853} & \centering \textbf{0.947} & \centerline{\textbf{0.979}}\\[-2.5ex]
		 Ours (diff-BA) & \centering CS+K & \centering none & \centering \textbf{0.107} & \centering \textbf{0.989} & \centering \textbf{4.868} & \centering \textbf{0.195} & \centering \textbf{0.867} & \centering \textbf{0.955} & \centerline{\textbf{0.981}}\\[-2ex]\hline
	\end{tabular}
	\caption{Comparison of quantitative results of single-view depth estimation results on KITTI Eigen \textit{et al.} \cite{eigen2014depth} test split. For training ``K'' refers to KITTI dataset \cite{geiger2013vision} (Eigen split) and ``CS'' refers to Cityscapes dataset \cite{cordts2016cityscapes}. For supervision depth means ground truth is used and pose means calibrated camera pose is used. All results are evaluated with depth capped at 80m except Garg \textit{et al.} \cite{garg2016unsupervised} which is capped 50m. ``$^{*}$'' is updated results from Zhou \textit{et al.} \cite{zhou2017unsupervised}.}
	\label{table1}
\end{table*}

\subsubsection{Cityscapes Dataset}
\begin{figure}[ht]
	\begin{center}
		\includegraphics[width=0.48\textwidth]{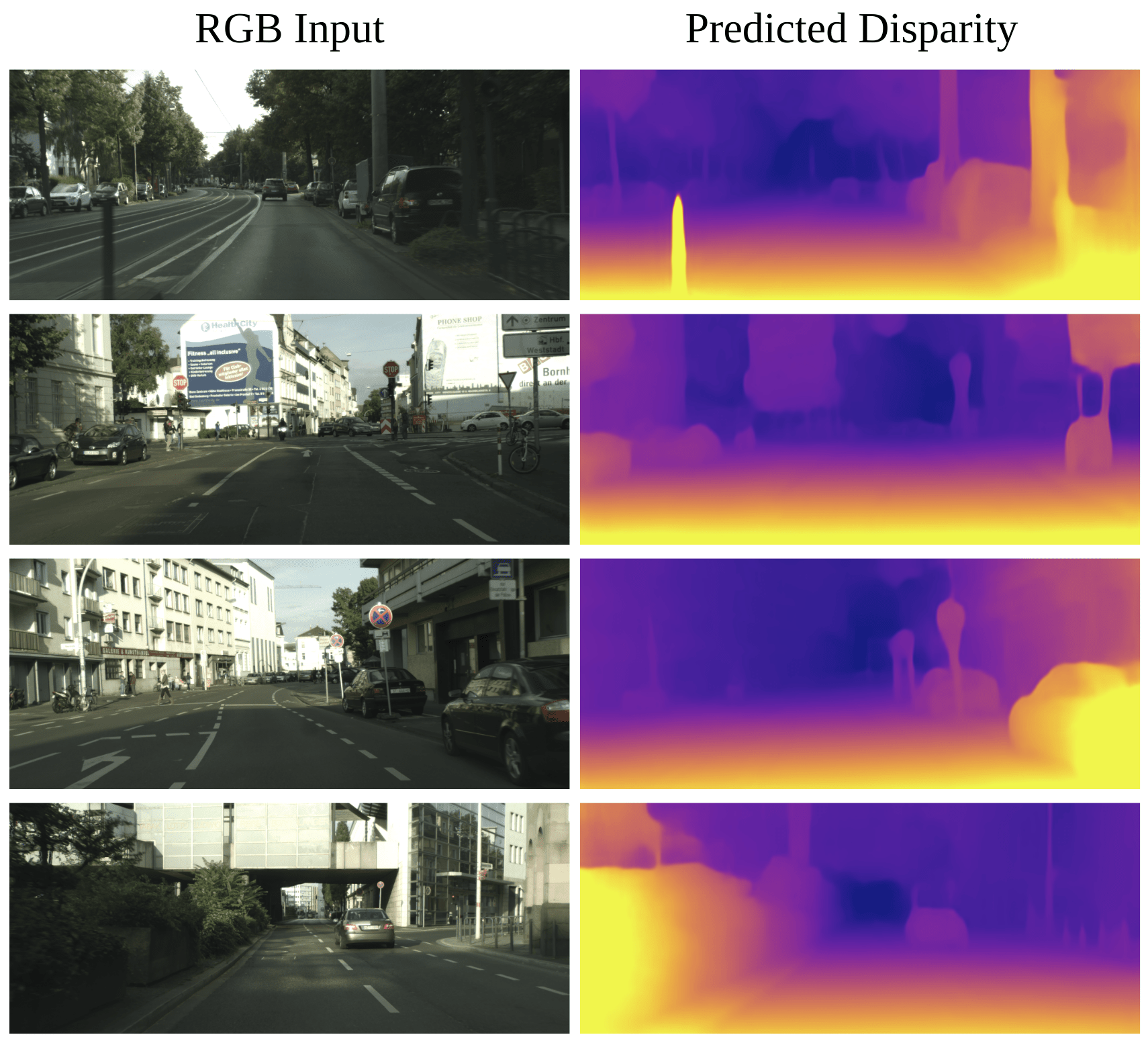}
	\end{center}
	\caption{Example qualitative single-view depth estimation results on the Cityscapes \cite{cordts2016cityscapes} test set. The first column is input RGB images and the second row is our depth predictions (visualized as disparities). Best viewed in color.}
	\label{fig:cityquality}
\end{figure}

Cityscapes \cite{cordts2016cityscapes} is a large-scale dataset that contains stereo video sequences recorded in street scenes from 50 different cities. As pointed out in \cite{godard2017unsupervised,zhou2017unsupervised}, pre-training the system on the larger Cityscapes and then fine-tuning on KITTI leads to performance improvements. Here we conduct a similar experiment.  We form the training sequences the same as in KITTI by setting the length of image sequence to 3, with the central frame being the target view and $\pm1$ frames being the source views. We first only train on Cityscapes and evaluate on the Cityscapes test set. We show some randomly sampled depth estimation results in Figure \ref{fig:cityquality}. One can see that our method still generates good depth prediction despite Cityscapes contains more dynamic scenes (i.e. more moving objects) and preserves thin structures like poles and trees. Next we use the Cityscapes trained model to fine-tune on KITTI and show the quantitative results in Table \ref{table1}. From Table \ref{table1} we can see that our model consistently out-performs all previous unsupervised approaches that are based on monocular video inputs and methods trained with rectified stereo pairs, which demonstrates that the differentiable BA layer better learns suitable features for Structure-from-Motion (SfM) tasks such as monocular depth estimation.

\subsection{Pose Estimation}
We evaluate the performance of our system on pose estimation on the official KITTI odometry split \cite{geiger2012we}, which contains 11 driving sequences with ground truth odometry obtained through the IMU/GPS readings. We use sequences from 00 to 08 for training and 09-10 are used for testing. In pose evaluation we set the length of image sequences to 5, with the central frame being the target view and $\pm2$ views being source views. We compare our pose estimation with two variants of monocular ORB-SLAM \cite{mur2015orb} as in \cite{zhou2017unsupervised}: 1) \textit{ORB-SLAM (full)}, which recovers camera poses using all of the driving sequences (i.e. with loop closure and re-localization), and 2) \textit{ORB-SLAM (short)}, which runs on 5-frame long tracklets that is same as our input setting. We ignore the first 9 frames of sequence 09 and the first 30 frames of sequence 10 for which ORB-SLAM fails to bootstrap with reliable camera poses due to large rotations and lack of good features. We report the Absolute Trajectory Error (ATE) \cite{sturm2012benchmark} which measures difference between the points of true and estimated trajectory. Given estimated trajectory $\bm{\mathrm{P}}_{1:n}$ and ground truth trajectory $\bm{\mathrm{Q}}_{1:n}$, the ATE at time step $i$ is computed as 
\begin{equation}
\bm{\mathrm{F}}_i = \bm{\mathrm{Q}}_i^{-1}\bm{\mathrm{S}}\bm{\mathrm{P}}_i,
\end{equation}
where $\bm{\mathrm{S}}$ is the rigid-body transformation corresponding to the least-square solution that maps $\bm{\mathrm{P}}_{1:n}$ to $\bm{\mathrm{Q}}_{1:n}$ which can be obtained using the method in \cite{horn1987closed}. Then the final result is obtained by computing the root mean squared error over all time steps
\begin{equation}
    \text{RMSE}_{\bm{\mathrm{F}}_{1:n}} = \Bigg{(}\frac{1}{n}\sum_{i=1}||t(\bm{\mathrm{F}}_i)||^2\Bigg{)}^{\frac{1}{2}},
\end{equation}
where $t(\cdot)$ is the translational component. As shown in Table \ref{table2}, our method out-performs ORB-SLAM (short) by a large margin and also gains advantage to the full ORB-SLAM system.
\begin{table}[H]
    \centering
    \footnotesize
    \begin{tabular}{p{2.8cm}p{2.cm}p{2.cm}} 
    \hline\hline 
    Method & \centerline{Seq. 09} & \centerline{Seq.10}\\[-1.5ex]\hline 
    ORB-SLAM (full) \cite{mur2015orb} & \centerline{0.014$\pm$0.008} & \centerline{0.012$\pm$0.011} \\[-1.5ex]
    ORB-SLAM (short) \cite{mur2015orb} & \centerline{0.064$\pm$0.141} & \centerline{0.064$\pm$0.130} \\[-1.5ex]
    Zhou \textit{et al.} \cite{zhou2017unsupervised} & \centerline{0.021$\pm$0.017} & \centerline{0.020$\pm$0.015}\\[-1.5ex]
    Zhou \textit{et al.} \cite{zhou2017unsupervised}$^{*}$ & \centerline{0.016$\pm0.009$} & \centerline{0.013$\pm 0.009$}\\[-1.5ex] 
    GeoNet \cite{yin2018geonet} & \centerline{\bm{$0.012\pm 0.007$}} & \centerline{0.012$\pm 0.009$}\\[-1.5ex]\hline
    Ours & \centerline{\bm{$0.012\pm 0.007$}} & \centerline{\bm{$0.011\pm 0.007$}} \\[-1.5ex]\hline
    \end{tabular}
    \caption{Absolute Trajectory Error (ATE) of pose estimation evaluated on the KITTI odometry split sequences 09-10. ``$^{*}$'' is updated results from Zhou \textit{et al.} \cite{zhou2017unsupervised}.}
    \label{table2} 
\end{table}
\section{Conclusion}
We have presented a novel self-supervised learning framework that tackles the tasks of scene depth and ego-motion estimation using unlabeled monocular videos as input. Our system jointly optimizes depth and camera pose through a differentiable bundle adjustment layer that enforces multi-view geometry constraints between 3D scene structures and camera motion. The whole pipeline is end-to-end trainable. Superior performance on large-scale public benchmarks proves the success of our method.

However there are still some challenges that are left to be addressed. First our current framework does not explicitly model the non-rigid parts of 3D scenes (e.g. moving objects), which is crucial for 3D scene understanding. Another issue is that although our system only requires unlabeled monocular videos, the camera intrinsics still need to be known. This prevents using Internet videos that are arguably the most vast source. We plan to address these challenges in our future work. Also as pointed out in \cite{zhou2017unsupervised} since depth maps are simplified representation of the underlying 3D scene, it would also be interesting to extend our current system to learn full 3D volumetric representations.

{\small
\bibliographystyle{ieee}
\bibliography{ref}
}

\end{document}